\pdfoutput=1
\documentclass[11pt]{article}
\usepackage{acl}   
\usepackage{times,latexsym}
\usepackage[T1]{fontenc}
\usepackage[utf8]{inputenc}
\usepackage{booktabs,graphicx,amsmath,amssymb}
\usepackage{eurosym}
\title{Grounded, Compute-Efficient LLM Policy Agents for Energy-Poverty Equity in Physically-Constrained Peer-to-Peer Energy Markets}
\author{Kunal Jadhav\thanks{\ \ Equal contribution.} \\ Arizona State University \\ \texttt{kjadha12@asu.edu} \\
  \And Siddhesh More\footnotemark[1] \\ Arizona State University \\ \texttt{smore123@asu.edu}}
\date{}

\begin{document}
\renewcommand{\thefootnote}{\fnsymbol{footnote}}
\maketitle
\renewcommand{\thefootnote}{\arabic{footnote}}\setcounter{footnote}{0}

\begin{abstract}
Energy poverty is nearly absent from NLP-for-social-good, and the little existing work is either static retrieval/QA or relies on carbon-intensive cloud LLMs, a self-defeating ``computational irony'' for a humanitarian setting. We present \textbf{EqGrid}, a closed-loop simulation in which a low-frequency, open-weight LLM \emph{policy agent} sets price and carbon bounds and targeted subsidies over a community of empirically-grounded household personas, while high-frequency multi-agent RL traders clear a continuous double auction constrained by a physical distribution grid (IEEE-33-bus with Dynamic Operating Envelopes). Our contribution is threefold and directly answers the workshop's special theme of \emph{how to measure the social impact of AI}: (i) \emph{grounded} personas (region-matched socio-demographics) whose load curves are checked for shape and level realism against real smart-meter data; (ii) formal energy-poverty equity metrics (Energy Burden, Gini of EB, LIHC) showing the intervention reduces burden inequality without raising net grid cost; and (iii) a \emph{compute-efficiency frontier} that measures how much equity performance survives compressing the policy agent from a 235B teacher down to a sub-1B model deployable on a laptop, in estimated energy/carbon per decision. A decoupled-safety design (the LLM sets bounds; a validate-and-project grid gate executes) yields \textbf{zero} grid-constraint violations versus 55 under direct LLM control. We will release code and configs. On energy-poverty equity, the LLM policy lowers the Gini of energy burden to 0.305 (from 0.351) and mean burden by 28\% while cutting cost (outperforming a tuned rule baseline), and a 3B-active model retains 95\% of the benefit at $\approx$9$\times$ lower inference energy than the teacher, with even a 0.8B on-device model retaining 92\% at $\approx$24$\times$ lower energy.
\end{abstract}

\section{Introduction}
Energy poverty, the inability to afford adequate domestic energy services, affects roughly
one in ten households in the European Union and is a direct concern of UN Sustainable
Development Goals 7 (affordable clean energy) and 10 (reduced inequalities). Yet it is nearly
absent from NLP for social good: recent surveys find poverty and energy among the least-studied
domains in the ACL community \citep{nlp4sg2025,lefevre2025}. The little existing work is either
static retrieval/QA over energy documents \citep{weqa2024} or, increasingly, relies on large
cloud LLMs whose inference carbon footprint \citep{howhungry2025} undercuts the very
sustainability goal it serves, a ``computational irony'' especially acute for humanitarian,
resource-constrained deployments.

A parallel line of work couples LLMs with multi-agent reinforcement learning (MARL) to run
peer-to-peer (P2P) energy markets \citep{fairmarket2025,expert2025}, but each system covers only
a slice of the problem. Fairness-shaping approaches optimise a \emph{generic} market-fairness
score rather than energy-poverty equity; equity-aware markets omit any language model
\citep{equityaware2025}; LLM-for-grid systems address safety but not markets or households
\citep{gridagent2025,rl2}; and the socioeconomic LLM-agent literature grounds personas in real
data but never touches energy or physical grids \citep{econagent2024,park2024}. Even
\citet{qiu2021}, the standard MARL double-auction baseline, explicitly defers the physical
network constraints to future work. EqGrid integrates these separate lines into a single framework
for energy-poverty mitigation: to the best of our knowledge, it is the first system to jointly combine
grounded energy-poverty personas, a compute-efficient LLM policy layer, MARL trading, and a
physically constrained grid while evaluating both energy-poverty equity \emph{and} the AI system's
own carbon cost. The contribution therefore lies in the integrated system and the measurement
framework rather than in any individual algorithmic component.

We present \textbf{EqGrid}, which closes exactly this gap and, in doing so, directly engages
this year's workshop theme of \emph{how to measure the social impact of AI}. Our contributions:
\begin{itemize}
\itemsep0em
\item A closed-loop framework in which a low-frequency, open-weight LLM \emph{policy agent}
sets price/carbon bounds and targeted subsidies over empirically grounded (EU-SILC) household
personas, while MARL traders clear a double auction under an IEEE-33-bus grid with Dynamic
Operating Envelopes.
\item A \textbf{measurement suite for social impact}: energy burden, Gini of energy burden,
LIHC prevalence, and a carbon-adjusted Social Return on Investment, with results showing the
LLM policy reduces burden inequality (Gini 0.305 vs 0.351) and cost while
beating a transparent rule baseline.
\item A \textbf{compute-efficiency frontier} tracing how much of this social benefit survives
compressing the policy agent from a 235B teacher to a sub-1B laptop-deployable model: a 0.8B
open model retains 92\% of the equity gain at $\approx$24$\times$ less energy per decision,
making the policy layer's sustainability cost an explicit, quantified trade-off.
\item A decoupled-safety design (the LLM sets bounds; a validate-and-project grid gate executes)
that yields \textbf{zero} grid violations versus 55 under direct LLM control.
\end{itemize}
We will release all code, configurations, and the measurement harness upon publication.

\section{Related Work}
\paragraph{NLP for energy and social good.} Surveys of the ACL Anthology identify poverty and
energy as among the least-addressed social-good domains \citep{nlp4sg2025,lefevre2025}. Existing
energy-domain NLP is largely informational, e.g., retrieval-augmented QA over technical documents
\citep{weqa2024}, rather than tied to a physical intervention. We move from answering questions
\emph{about} energy to simulating interventions \emph{on} energy poverty.

\paragraph{LLM socioeconomic agents.} LLMs can simulate economic agents \citep{homosilicus2023}
and macroeconomic households \citep{econagent2024}, and grounding agents in real interviews/surveys
improves population-level fidelity \citep{park2024}, though reliability is distributional rather
than individual \citep{digitalpersonas2026}. \citet{llmeconomist2025} place an LLM planner over
census-conditioned agents to design tax policy. We adapt this planner-over-population design to a
\emph{physically constrained energy market} with energy-poverty equity and edge efficiency.

\paragraph{MARL and LLM-guided P2P markets.} Double-auction P2P trading is commonly solved with
MARL \citep{qiu2021}; recent work adds LLM guidance, whether fairness shaping \citep{fairmarket2025},
expert-imitation workflows \citep{expert2025}, or explainable equity without an LLM
\citep{equityaware2025}. These optimise generic fairness or cost and omit grounded personas,
edge efficiency, and, as \citet{qiu2021} note, physical network constraints. They also couple the
language model to the \emph{fast} trading loop, as a per-episode fairness critic
\citep{fairmarket2025} or an expert-imitation signal \citep{expert2025}. Our LLM instead acts
\emph{ex ante} as a low-frequency policy layer that sets the market's price and carbon bounds and
targeted subsidies, which the MARL traders then operate within. This forward, low-frequency role is
also what makes the compute-efficiency question meaningful: the policy model runs only a handful of
times per day rather than inside every trade, so a small on-device backbone is viable. LLM-for-grid
systems contribute an LLM-reasoning-plus-solver safety pattern \citep{gridagent2025,rl2} that we
adopt for our decoupled grid gate.

\paragraph{Green AI.} Inference energy varies by orders of magnitude across models
\citep{howhungry2025,sustainablenlp2025}, and quantization/compression reduce it, with hardware
caveats. We use this literature both to \emph{measure} our policy agent's footprint and to argue
for small, open-weight, on-device backbones. To our knowledge, no prior energy-poverty system
reports the carbon cost of its own AI. A full literature matrix is provided in the appendix.

\section{The EqGrid Framework}
\label{sec:method}
\begin{figure}[t]\centering
\includegraphics[width=\linewidth]{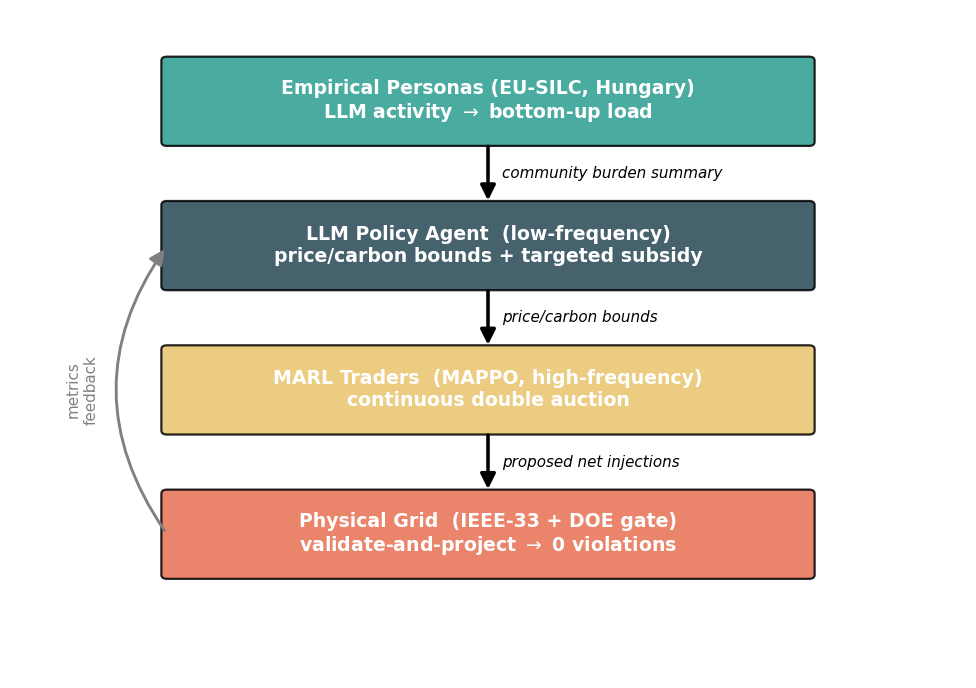}
\caption{The EqGrid closed loop. The LLM policy agent sets economic bounds only; a
validate-and-project grid gate executes trades safely. Persona grounding, MARL trading, and the
physical grid close the loop.}
\label{fig:arch}
\end{figure}

EqGrid couples four components in a closed loop (Figure~\ref{fig:arch}): (i) empirically
grounded household personas; (ii) a low-frequency, open-weight LLM \emph{policy agent} that
sets market bounds; (iii) high-frequency MARL \emph{traders} that clear a continuous double
auction within those bounds; and (iv) a physical distribution-grid layer that enforces
safety. By design, the LLM never executes physical actions; it only sets bounds, so grid
operation stays deterministic and safe (\S\ref{sec:safety}).

\subsection{Empirically grounded personas}
Each household $i$ is instantiated from published EU-SILC marginals for Hungary: household
size, equivalised income, dwelling energy-efficiency class, heating type, and the two EU
energy-poverty indicators (utility-bill arrears; inability to keep the home adequately warm).
An LLM converts each persona into a 24-hour activity-intensity profile via budget- and
comfort-aware chain-of-thought; a deterministic bottom-up appliance-and-heating model then
maps activities to an hourly load curve, with fuel-poor households curtailing heating
(capturing ``hidden'' energy poverty). Prosumers additionally receive a winter PV generation
curve. We compare three grounding conditions, namely \textsc{grounded} (full attributes),
\textsc{synthetic} (randomised attributes), and \textsc{demographic} (size+income only,
following the demographics-only baseline of \citealp{park2024}), and check the resulting
load distributions for shape and level realism against real smart-meter data (\S\ref{sec:exp}).

\subsection{LLM policy agent}
Every policy period (6h) the agent observes an anonymised community summary (the distribution
of energy burdens, the Gini of energy burden, the share of high-burden and fuel-poor
households, and current tariffs) and returns a JSON policy: a price ceiling and floor, a
low-income subsidy weight, and a per-household carbon allowance. The prompt instructs the agent
to reduce energy-poverty inequality without raising net grid cost, cap peak prices, and keep
floors above the feed-in tariff so exporters stay viable. We compare the LLM agent against a
transparent \textsc{rule} baseline (tighten the ceiling as inequality rises; scale subsidy to
the arrears rate) and a \textsc{none} baseline (no intervention), isolating the value of LLM
reasoning.

\subsection{MARL traders and market}
Each household is a trading agent controlling a home battery and its bid price. Agents are
trained with a compact MAPPO (shared-parameter actor, centralized critic on a public market
summary), echoing the order-book-fed critic of \citet{qiu2021} under
centralized-training/decentralized-execution. Each hour agents submit price--quantity orders
bounded to the policy band; a continuous double auction clears at the mid-price, and unmatched
volume settles with the utility at time-of-use / feed-in tariffs.

The LLM's 6-hourly bounds enter the trader loop without any real-time retraining, in two ways.
First, the current period's price ceiling and floor are part of each agent's observation (and the
centralized critic's state), so the policy is conditioned on the active band. Second, each agent
emits a continuous bid variable in $(0,1)$ that is affine-rescaled into the current
$[\text{floor},\text{ceiling}]$ interval at order-submission time, so bids automatically track a
change in bounds; the subsidy weight enters the reward directly, reshaping incentives for low-income
households. A single MAPPO policy is thus trained once over the full horizon with the time-varying
bounds in place, rather than being re-optimised whenever the LLM updates the band.

\subsection{Physical grid layer and decoupled safety}
\label{sec:safety}
Cleared net injections are aggregated to an IEEE-33-bus radial feeder (each bus is a low-voltage
load pocket) and checked by a Newton--Raphson power flow (pandapower). The \emph{Dynamic
Operating Envelope} is the set of injections that keep all bus voltages in $[0.95,1.05]$ pu and
all lines below their thermal limit. A validate-and-project gate enforces it: if the proposed
injections are infeasible, we project them onto the feasible set by bisection on a global
curtailment factor, guaranteeing zero violations while curtailing as little as possible. The
LLM-direct-control ablation removes this gate, exposing violations.

\subsection{Compute-efficiency frontier}
Because the policy agent is the only LLM in the loop and runs at low frequency, we can vary its
backbone along a ladder from a 235B-parameter teacher down to a sub-1B model that runs locally
on a laptop, holding persona generation fixed. For each rung we measure the equity outcome and
the energy/carbon per policy decision, tracing how much social-impact benefit survives
compression. This operationalises the workshop's ``how to measure social impact'' theme
jointly with the carbon cost of the AI itself.

\section{Measuring Social Impact: Metrics}
Answering the workshop's special theme requires measuring both the social benefit and the AI's
own cost. We adopt five measures.

\paragraph{Energy Burden (EB).} For household $i$, $EB_i = C^{\text{energy}}_i / I_i$, the ratio
of net annual energy cost (grid imports net of local trades and PV) to income.

\paragraph{Gini of Energy Burden.} Community burden inequality,
$G_{EB} = \frac{\sum_i \sum_j |EB_i - EB_j|}{2 N^2 \overline{EB}}$; a successful policy lowers
$G_{EB}$ \emph{without} raising net grid cost.

\paragraph{LIHC prevalence.} The Low-Income-High-Cost count: households below median income whose
energy share exceeds 10\%. We also report the 90th-percentile burden to track the vulnerable tail
and guard against ``hidden'' energy poverty from under-consumption \citep{energyequitygap2022}.

\paragraph{Carbon-adjusted SROI.}
$\text{SROI}_{\text{net}} = (\Delta V_{\text{social}} - C_{\text{carbon}})/I_{\text{initial}}$,
where $\Delta V_{\text{social}}$ is monetised bill savings and avoided-outage value, and
$C_{\text{carbon}}$ is the monetised carbon of LLM inference. This explicitly nets the AI's
footprint against its social benefit.

\paragraph{Energy/carbon per decision.} For each policy decision we log the generated token count
and estimate energy with a transparent proxy (J/token scaled by \emph{active} parameters, calibrated
to \citealp{howhungry2025}) and $\text{CO}_2$e at the regional grid intensity. These are proxy
estimates from measured token counts, not on-device power measurements; direct hardware profiling of
the small models is future work (see Limitations). This measure drives the compute-efficiency
frontier (\S\ref{sec:results}).

\section{Experiments}
\label{sec:exp}
\paragraph{Setup.} We simulate a Budapest residential community on an IEEE-33-bus feeder (each
bus a low-voltage load pocket), over a 24-hour winter horizon with hourly double-auction rounds
and a 6-hourly policy period. Personas are drawn from published EU-SILC marginals for Hungary
(income, household size, dwelling efficiency, heating type, and the two EU energy-poverty
indicators). Load curves are generated by LLM activity profiles plus a bottom-up appliance/heating
model and checked for shape and level realism against a real smart-meter reference
\citep{uci_household}, which is out of region and period (see Limitations). MARL traders use MAPPO;
the policy agent is an open-weight instruction-tuned model served over an OpenAI-compatible API.
Exact public checkpoints and licenses for all six ladder models are listed in
Appendix~\ref{app:models}.

\paragraph{Protocol and statistics.} We run five seeds and report the mean $\pm$ standard deviation.
Because a seed fixes the persona population, the policy conditions are \emph{paired} across seeds, so
we assess significance with paired $t$-tests and Wilcoxon signed-rank tests on the per-seed
differences and report 95\% confidence intervals and the paired effect size $d_z$. With five seeds
these tests are conservative (the smallest attainable one-sided Wilcoxon $p$ is $0.03$), so we treat
results as significant only when both tests and the CI agree.

\paragraph{Conditions.} (i) \emph{Policy} (H2): \textsc{none} / \textsc{rule} / \textsc{llm}.
(ii) \emph{Grounding ablation}: \textsc{grounded} vs \textsc{synthetic} personas under the LLM
policy, measuring subsidy-to-burden targeting alignment. (iii) \emph{Compute ladder} (H3): the
policy backbone varied from a 235B teacher to a sub-1B laptop model, holding persona generation
fixed. (iv) \emph{Safety} (H4): DOE gate on vs off (LLM-direct-control).

\paragraph{Baselines.} We compare against the no-policy MARL market (the DA-MADDPG-style setting
of \citealp{qiu2021}), a transparent rule-based policy, and, for energy, a large cloud model as
the ``carbon-intensive'' reference.

\section{Results}
\label{sec:results}

\paragraph{Equity (H2).} Table~\ref{tab:h2} reports means $\pm$ standard deviation over five seeds;
the LLM policy agent improves on both the no-policy and rule baselines on every equity and cost
measure. It attains the lowest Gini of energy burden (0.305, vs.\ 0.330 for the rule and 0.351 for
no policy), the lowest mean burden (0.247$\rightarrow$0.177), the fewest low-income-high-cost
households (14.8$\rightarrow$12.4), and the lowest daily cost (\euro148.0$\rightarrow$\euro120.3),
while achieving the highest carbon-adjusted SROI (101.1 vs.\ 45.0 for the rule; $\approx$0 for no
policy). We assess significance with paired tests across seeds, where each seed fixes the persona
population. Against no policy, the LLM significantly lowers both the Gini of burden (mean reduction
0.046, 95\% CI $[0.013, 0.079]$, paired $t$-test $p{=}0.009$, Wilcoxon $p{=}0.03$, $d_z{=}1.7$) and
daily cost (\euro27.7, CI $[24.9, 30.4]$, $p{<}0.001$). Against the \emph{tuned rule} it
significantly reduces cost (\euro15.4, CI $[9.3, 21.4]$, $p{=}0.001$) and raises SROI ($+56$, CI
$[34, 78]$, $p{=}0.001$); its additional reduction of the Gini over the rule is directional but not
robustly significant at five seeds (0.025, CI $[-0.003, 0.052]$, $t$-test $p{=}0.03$, Wilcoxon
$p{=}0.06$). The rule comparison isolates the value of \emph{reasoning}: it indicates that context-sensitive
policy generation provides benefit beyond a transparent hand-designed rule, with the evidence
strongest on cost, SROI, and equity relative to no policy, and only suggestive for equity relative
to the rule.

\begin{table}[t]\centering\small
\resizebox{\columnwidth}{!}{%
\setlength{\tabcolsep}{4pt}%
\begin{tabular}{lrrrr}
\toprule
Policy & Gini(EB) & mean EB & LIHC & bill (\euro/day) \\
\midrule
none & 0.351\,{\scriptsize$\pm$0.017} & 0.247\,{\scriptsize$\pm$0.043} & 14.8 & 148.0\,{\scriptsize$\pm$12.6} \\
rule & 0.330\,{\scriptsize$\pm$0.012} & 0.217\,{\scriptsize$\pm$0.031} & 14.2 & 135.7\,{\scriptsize$\pm$9.4} \\
llm & 0.305\,{\scriptsize$\pm$0.023} & 0.177\,{\scriptsize$\pm$0.030} & 12.4 & 120.3\,{\scriptsize$\pm$10.8} \\
\bottomrule
\end{tabular}}

\caption{Equity by policy (mean $\pm$ std over 5 seeds). Lower Gini/EB/LIHC and bill are better.}
\label{tab:h2}
\end{table}

\paragraph{Grounding ablation.} Empirical grounding yields marginally better subsidy-to-burden
targeting (alignment 0.79$\pm$0.05 vs.\ 0.70$\pm$0.09 for synthetic personas), but the paired
difference is \emph{not} statistically significant at five seeds (0.08, 95\% CI $[-0.04, 0.21]$,
$p{=}0.06$). We are deliberately cautious here: the
two conditions induce \emph{different} synthetic populations (grounded personas reproduce the
realistic right-skewed EU-SILC income and burden structure, whereas ungrounded personas are more
uniform), so their absolute equity levels are not directly comparable, and grounding is not the
source of a large headline gain. We therefore treat empirical grounding as a construct-validity and
realism choice: it reproduces the socioeconomic structure that defines \emph{who} is energy-poor,
and the resulting load curves are consistent in diurnal shape and level with a real smart-meter
reference (Appendix), subject to that reference being out of region and period (see Limitations). Establishing
a decisive grounded-vs-synthetic effect would require restricted EU-SILC microdata and a
multi-household load benchmark, which we flag as future work.

\begin{figure}[t]\centering
\includegraphics[width=\linewidth]{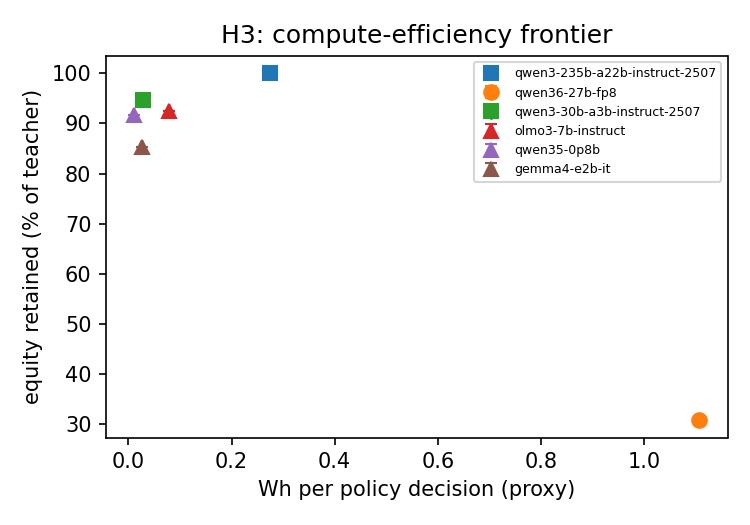}
\caption{Equity retained (\% of teacher) vs energy per decision. Squares: MoE; triangles:
laptop-deployable.}
\label{fig:frontier}
\end{figure}

\paragraph{Compute-efficiency frontier (H3).} Table~\ref{tab:h3} and Figure~\ref{fig:frontier}
trace equity retained against energy per policy decision across the model ladder, revealing a clear
sweet spot that is \emph{not} the largest model. The 235B teacher sets the reference at 0.27~Wh per
decision. A small Mixture-of-Experts agent (\texttt{qwen3-30b-a3b}, 3B active) retains \textbf{95\%}
of its equity gain at \textbf{0.029~Wh}, a $\approx$9$\times$ energy reduction. The
social benefit survives all the way down to \emph{laptop-deployable} open models: \texttt{olmo3-7b}
(fully open) retains 92\% at 0.078~Wh, and even a \textbf{0.8B} agent (\texttt{qwen35-0p8b}) retains
\textbf{92\%} of the teacher's equity gain at just \textbf{0.011~Wh}, a $\approx$24$\times$ energy
reduction, with \texttt{gemma4-e2b} (2B) at 85\%/0.026~Wh. This retention is statistically real: over
five seeds the teacher and all four small/MoE models fall within about one standard deviation of one
another on the Gini (0.302--0.310, std $\approx$0.02--0.03; Table~\ref{tab:h3}), i.e.\ they are
statistically indistinguishable, whereas no policy (0.351) and the dense 27B (0.336) sit clearly
outside that band. Compression to a laptop-scale model therefore costs no statistically detectable
equity. The one model that
\emph{does} collapse is a \emph{larger dense reasoning} model (\texttt{qwen36-27b}): it is
simultaneously the worst on equity (retaining only 31\%) and by far the most energy-hungry
(1.11~Wh, $\approx$4$\times$ the teacher and $\approx$100$\times$ the 0.8B agent).

An alternative explanation for this collapse is that the reasoning model is merely
\emph{under-budgeted}, so we address that possibility directly. First, the mechanism is explicit in
our logs: this model spends its entire generation budget on chain-of-thought at every step (hitting
the length limit on 100\% of calls), so on 90\% of decisions the output is truncated before a
closeable JSON object is emitted and the agent falls back to its default bounds.\footnote{The agent
parses the model's structured output tolerantly, reading the \texttt{reasoning\_content} channel when
a reasoning model leaves the primary content field empty. Empty-parse rates over our released
response cache are 0/20 for the teacher, \texttt{qwen35-0p8b} and \texttt{olmo3-7b}, 1/20 for
\texttt{gemma4-e2b}, and 18/20 for \texttt{qwen36-27b}.} Its 31\% therefore reflects
\emph{truncation-induced fallback} rather than poor policy reasoning, which is itself the
deployment-relevant failure: at a budget that keeps its energy cost within an order of magnitude of
the teacher's, the model does not return a usable decision at all. This failure mode is intrinsic
to deploying a reasoning-tuned model on a task that wants a short structured answer. Second, and decisively, the conclusion is \emph{robust to the budget
choice}: raising the token limit cannot help without raising energy, since energy scales with
generated tokens. Even under the most charitable counterfactual, in which a larger budget lets it
fully reason and match the teacher's 100\% equity, it would only move \emph{further right} on the frontier
(to several Wh per decision, an order of magnitude past the teacher and $\approx$1000$\times$ the
0.8B agent), remaining strictly Pareto-dominated by the small and MoE models. No token budget places
a dense reasoning model on the efficient frontier for this low-frequency control task. We therefore
frame this as a \emph{deployment} finding, namely that reasoning-heavy LLMs are the wrong tool for
cheap, high-cadence policy decisions, rather than a claim about the model's general capability.

The socially-best-per-joule policy agent therefore runs on commodity hardware, giving a
quantified argument for small, open, on-device models over larger,
reasoning-heavy LLMs on this low-frequency task.

\begin{table}[t]\centering\small
\resizebox{\columnwidth}{!}{%
\setlength{\tabcolsep}{4pt}%
\begin{tabular}{lrrrrr}
\toprule
Model & act.B & Gini(EB) & eq.ret\% & Wh/dec & \$ \\
\midrule
qwen3-235b-a22b-instruct-2507 & 22.0 & 0.302\,{\scriptsize$\pm$0.026} & 100 & 0.2745 & 0.0764 \\
qwen36-27b-fp8 & 27.0 & 0.336\,{\scriptsize$\pm$0.022} & 31 & 1.1071 & 0.7085 \\
qwen3-30b-a3b-instruct-2507 & 3.0 & 0.305\,{\scriptsize$\pm$0.023} & 95 & 0.0292 & 0.1715 \\
olmo3-7b-instruct & 7.0 & 0.306\,{\scriptsize$\pm$0.023} & 92 & 0.0784 & 0.0605 \\
qwen35-0p8b & 0.8 & 0.306\,{\scriptsize$\pm$0.028} & 92 & 0.0113 & 0.0611 \\
gemma4-e2b-it & 2.0 & 0.310\,{\scriptsize$\pm$0.017} & 85 & 0.0262 & 0.1133 \\
\bottomrule
\end{tabular}}

\caption{Compute-efficiency frontier: Gini (mean $\pm$ std over 5 seeds), equity retained (\% of
teacher gain), and energy/cost per policy decision.}
\label{tab:h3}
\end{table}

\paragraph{Safety (H4).} The DOE validate-and-project gate yields \textbf{0} voltage/line
violations (128 bus-hours curtailed), whereas removing it
(LLM-direct-control) produces 55 violations. The LLM never needs to be trusted
with physical control.

\section{Limitations}
Our study has several limitations, stated plainly. \textbf{(1) Persona grounding.} We ground
personas in \emph{published} EU-SILC marginal distributions for Hungary rather than restricted
microdata (which requires a multi-week Eurostat application); we therefore reproduce marginal and
correlational structure, not a specific microdata sample. \textbf{(2) Load-validation reference is
out of region and period.} Our real smart-meter reference is the UCI single-household dataset
\citep{uci_household}, which records one dwelling in France over 2006--2010, whereas our personas
represent an aggregate Budapest (Hungary) community. We therefore use it only as a coarse sanity
check on diurnal \emph{shape} and consumption \emph{level} plausibility, not as a region- or
period-matched validation of Hungarian load. A proper distributional validation would use an open
\emph{multi-household} benchmark closer to the target setting, such as Low Carbon London (half-hourly
readings for over 5{,}000 households); substituting or supplementing our reference with such a
dataset is the immediate next step, and we expect a robust grounding claim to rest on
multi-household distributional distance rather than a single-home comparison. Grounding improves
\emph{targeting} (who is helped) more than it changes aggregate load realism, and we frame it
accordingly. \textbf{(3) Scale.} We simulate a single feeder over a
representative winter day; larger networks, seasonal variation, and more households remain future
work. \textbf{(4) Energy accounting is a proxy, not a hardware measurement.} All per-decision energy
figures, including those for the small ``laptop-deployable'' models, are active-parameter-based
estimates from measured token counts, calibrated to the literature; we did not meter any model on
physical hardware in these experiments, and on-device profiling (e.g., Apple Silicon power counters
or CodeCarbon) is future work. We stress, however, that the frontier's
\emph{ordering} (in particular the dense reasoning model's dominated position) is driven by
\emph{measured} per-decision token counts and \emph{known} active-parameter counts; the only
estimated quantity is a global energy-per-token constant, which is a common multiplier across models
and therefore does not affect the relative comparison. The qualitative conclusion is thus robust to
the calibration. \textbf{(5) The subsidy is an unfunded transfer.} The targeted subsidy is applied as a discount to
below-median-income households without a corresponding levy, budget cap, or revenue-neutrality
constraint, so it redistributes burden without a modelled funding source. A consequence is that the
Gini of energy burden is largely determined by the magnitude of the single subsidy-weight scalar the
policy agent emits: regressing Gini on that scalar across our eight policy conditions gives
$R^2{=}0.985$. H2 and H3 should therefore be read as measuring whether a model emits a
well-calibrated policy scalar under a fixed transfer mechanism, and in particular whether that
calibration survives compression, rather than as evidence about the net welfare effect of a fundable
subsidy scheme; establishing the latter requires a revenue-neutral transfer and a re-run of the
model ladder, which we flag as the most important next step. \textbf{(6) Simulation, not deployment.}
Results are in-simulation; real deployment would require field validation with utilities and
affected communities, which we regard as essential before any real-world use.

\section*{Ethical and Societal Implications}
This work targets a positive social impact, reducing energy-poverty inequality, but operating on
vulnerable households demands care.

\paragraph{What ``positive impact'' means here.} We take positive impact to mean a measurable
reduction in the inequality of energy burden (Gini of EB, LIHC prevalence, the vulnerable tail)
that does not raise net cost and does not itself impose an outsized environmental cost. We make
this operational and auditable rather than rhetorical, and we net the carbon cost of our own AI
against its social benefit (SROI$_\text{net}$).

\paragraph{Fairness and disaggregated harm.} Algorithmic allocation can entrench bias. We enforce
and report a demographic-parity view across income groups and check that support reaches genuinely
high-burden households (the grounding ablation); a policy that lowered aggregate inequality while
harming a subgroup would be a failure. Multi-generational and low-efficiency-dwelling households
warrant particular scrutiny.

\paragraph{Safety: decoupling language models from physical control.} LLM hallucination is
dangerous in power systems. By construction the LLM only sets economic \emph{bounds}; a
deterministic power-flow gate projects all trades onto the safe operating envelope, yielding zero
grid violations. No physical action is ever taken on an LLM's unverified output.

\paragraph{Privacy.} Household consumption is sensitive. The design runs on small, open-weight,
on-device models, keeping data local and avoiding third-party cloud inference; the policy agent
sees only aggregate community statistics, never individual traces.

\paragraph{Risks of misuse and over-claiming.} The same market machinery could be tuned to extract
rather than protect; governance and transparency (we will release our code and an explainable rule
baseline) are mitigations. Our results are simulated, and we explicitly caution against deployment
without field validation and participation of the affected communities and utilities.

\bibliography{refs}

\appendix
\section{Model Identifiers and Licenses}
\label{app:models}
For reproducibility, Table~\ref{tab:models} gives the exact public checkpoint, parameter count,
and license for every model on the compute-efficiency ladder (\S\ref{sec:results}). All six are
open-weight and obtainable from public model hubs; the fully open OLMo-3 additionally releases its
training data and code. We served them unmodified through an OpenAI-compatible API, and the
\emph{active} parameter count of each drives our per-decision energy estimate.

\begin{table*}[t]\centering\footnotesize
\resizebox{\textwidth}{!}{%
\begin{tabular}{@{}p{5.7cm}p{6.6cm}p{1.8cm}p{1.8cm}p{2.8cm}@{}}
\toprule
\textbf{Served identifier} & \textbf{Public checkpoint} & \textbf{Params (total/active)} & \textbf{License} & \textbf{Openness} \\
\midrule
\texttt{qwen3-235b-a22b-instruct-2507} & \texttt{Qwen/Qwen3-235B-A22B-Instruct-2507} & 235B / 22B & Apache-2.0 & Open weights \\
\texttt{qwen36-27b-fp8} & \texttt{Qwen/Qwen3.6-27B-FP8} & 27B (FP8) & Apache-2.0 & Open weights \\
\texttt{qwen3-30b-a3b-instruct-2507} & \texttt{Qwen/Qwen3-30B-A3B-Instruct-2507} & 30B / 3B & Apache-2.0 & Open weights \\
\texttt{olmo3-7b-instruct} & \texttt{allenai/OLMo-3-7B-Instruct} & 7B & Apache-2.0 & Fully open (weights+data+code) \\
\texttt{qwen35-0p8b} & \texttt{Qwen/Qwen3.5-0.8B} & 0.8B & Apache-2.0 & Open weights \\
\texttt{gemma4-e2b-it} & \texttt{google/gemma-4-e2b-it} & 2B eff. & Gemma Terms & Open weights \\
\bottomrule
\end{tabular}}
\caption{Open-weight models on the compute-efficiency ladder, with public checkpoints and licenses.
Active parameters drive the per-decision energy estimate.}
\label{tab:models}
\end{table*}

\section{Literature Matrix}
\label{app:matrix}
Table~\ref{tab:matrix} positions EqGrid against the closest 2021--2026 work across the seven
threads it draws on. Each row shares one or more components with EqGrid; the rightmost column states
what it lacks relative to our integration. To our knowledge, EqGrid is the first to jointly couple
empirically grounded energy-poverty personas, a compute-efficient LLM policy, MARL trading, and a
physical grid, evaluated with formal energy-poverty equity and a measured account of the AI's own
carbon cost.

\begin{table*}[t]\centering\small
\renewcommand{\arraystretch}{1.15}
\begin{tabular}{@{}p{2.9cm}p{1.6cm}p{4.6cm}p{5.0cm}@{}}
\toprule
\textbf{Work} & \textbf{Venue} & \textbf{Method / key idea} & \textbf{Gap vs.\ EqGrid} \\
\midrule
\multicolumn{4}{@{}l}{\emph{NLP for energy / social good}}\\
\citet{weqa2024} & NLP4PI'24 & RAG QA benchmark over wind-energy documents & Static, informational; no agents, market, or physical intervention \\
\citet{nlp4sg2025} & EACL'26 & Survey of NLP4SG domains & Identifies poverty/energy as neglected; no method \\
\midrule
\multicolumn{4}{@{}l}{\emph{LLM socioeconomic / persona agents}}\\
\citet{econagent2024} & ACL'24 & LLM households in a macro loop & No empirical grounding, no energy/grid, no equity \\
\citet{park2024} & 2024 & Personas grounded in interviews/surveys & No energy or grid; population-level fidelity only \\
\citet{llmeconomist2025} & 2025 & LLM planner sets tax policy over census agents & Abstract economy; no grid physics, no edge efficiency \\
\citet{homosilicus2023} & 2023 & LLMs as economic agents & Foundational; validity caveats motivate grounding \\
\midrule
\multicolumn{4}{@{}l}{\emph{MARL \& LLM-guided P2P markets}}\\
\citet{qiu2021} & IJCAI'21 & DA-MADDPG double-auction traders & \textbf{Explicitly defers network constraints} (our contribution) \\
\citet{fairmarket2025} & 2025 & LLM critic shapes MARL with generic fairness & Not EP equity (EB/Gini/LIHC); no grounding, no edge cost, no IEEE-bus/DOE \\
\citet{expert2025} & IEEE TSG'26 & LLM expert imitation for MARL P2P & No equity across classes, no edge efficiency \\
\citet{equityaware2025} & 2025 & MILP+PPO+Shapley, Rawlsian equity & No LLM; no compute-efficiency axis \\
\midrule
\multicolumn{4}{@{}l}{\emph{LLM-for-grid safety}}\\
\citet{gridagent2025} & 2025 & LLM planner + power-flow solver + rollback & No market/persona/equity (we borrow the gate pattern) \\
\citet{rl2} & IEEE TSG'24 & LLM writes safe-RL penalty functions & Single-agent; no market or equity \\
\midrule
\multicolumn{4}{@{}l}{\emph{Green AI \& equity metrics}}\\
\citet{howhungry2025} & 2025 & Benchmarks energy/water/carbon of LLM inference & Measurement only; we apply it inside SROI \\
\citet{doep2p2024} & Appl.\ Energy'24 & DOE-embedded P2P-to-grid trading & Physical method; no LLM/persona/equity \\
\citet{energyequitygap2022} & Nat.\ Comms'22 & Energy equity gap (hidden EP) & Metric; motivates our LIHC + tail reporting \\
\bottomrule
\end{tabular}
\caption{Literature matrix. To our knowledge, EqGrid is the first to couple grounded EP personas + a
compute-efficient LLM policy + MARL trading + a physical (IEEE-33/DOE) grid, evaluated with
energy-poverty equity \emph{and} the AI's own carbon cost.}
\label{tab:matrix}
\end{table*}

\end{document}